%% file: acl2023.tex
\pdfoutput=1
\documentclass[11pt]{article}

\usepackage[]{ACL2023}

\usepackage{amsfonts}
\usepackage{amsmath}
\usepackage{amssymb}
\usepackage{booktabs}
\usepackage{cleveref}
\usepackage{etoolbox}
\usepackage{graphicx}
\usepackage{hyperref}
\usepackage{latexsym}
\usepackage{multirow}
\usepackage{pifont}
\usepackage{times}

\usepackage[T1]{fontenc}
\usepackage[utf8]{inputenc}

\usepackage{microtype}

\usepackage{inconsolata}

\newcommand{\MethodName}{Vec2Text}

\title{Text Embeddings Reveal (Almost) As Much As Text}
\author{John X. Morris, Volodymyr Kuleshov, Vitaly Shmatikov, Alexander M. Rush \\
Department of Computer Science \\
Cornell University}

\begin{document}
\maketitle

\begin{abstract}
How much private information do text embeddings reveal about the original text? We investigate the problem of embedding \textit{inversion}, reconstructing the full text represented in dense text embeddings. We frame the problem as controlled generation: generating text that, when reembedded, is close to a fixed point in latent space. We find that although a na\"ive model conditioned on the embedding performs poorly, a multi-step method that iteratively corrects and re-embeds text is able to recover $92\%$ of $32\text{-token}$ text inputs exactly. We train our model to decode text embeddings from two state-of-the-art embedding models, and also show that our model can recover important personal information (full names) from a dataset of clinical notes. \footnote{Our code is available on Github: \href{https://github.com/jxmorris12/vec2text}{github.com/jxmorris12/vec2text}.}
\end{abstract}

\section{Introduction}

\begin{figure*}[t]
    \centering
    \includegraphics[width=1.0\textwidth]{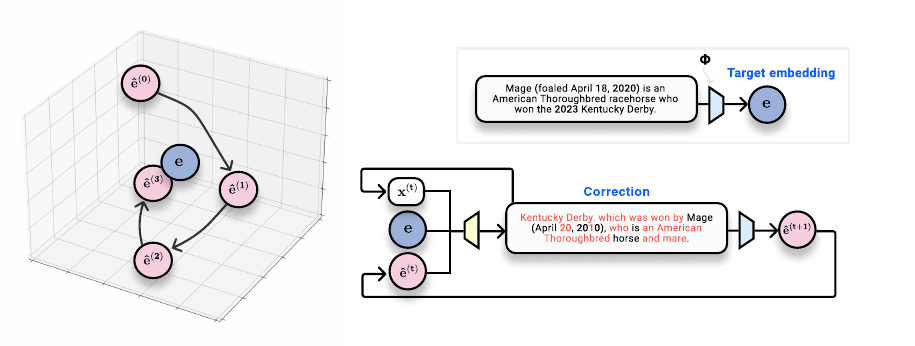}
    \caption{Overview of our method, \MethodName. Given access to a target embedding $e$ (blue) and query access to an embedding model $\phi$ (blue model), the system aims to iteratively generate (yellow model) hypotheses $\hat{e}$ (pink) to reach the target. Example input is a taken from a recent Wikipedia article (June 2023). \MethodName\ perfectly recovers this text from its embedding after $4$ rounds of correction.}
    % https://www.figma.com/file/31ZdB5bBLWx7YBxf1P3Ent/inverting-text-embeddings?type=design&node-id=0-1&t=qYxeddAJaw2HZfA4-0
    \label{fig:00_main}
\end{figure*}

Systems that utilize large language models (LLMs) often store auxiliary data in a vector database of dense embeddings \cite{borgeaud2022improving, yao2023react}. Users of these systems infuse knowledge into LLMs by inserting retrieved documents into the language model's prompt. Practitioners are turning to hosted vector  database services to execute embedding search efficiently at scale \cite{vdb1Pinecone,vdb2qdrant,vdb3vdaas,vdb4weaviate,harrison2023langchain}. 
In these databases, the data owner only sends \textit{embeddings} of text data \cite{le2014distributed,kiros2015skipthought} to the third party service, and never the text itself. The database server returns a search result as the index of the matching document on the client side.

Vector databases are increasingly popular, but privacy threats within them have not been comprehensively explored. Can the third party service to reproduce the initial text, given its embedding? Neural networks are in general non-trivial or even impossible to invert exactly. Furthermore, when querying a neural network through the internet, we may not have access to the model weights or gradients at all.

Still, given input-output pairs from a network, it is often possible to approximate the network's inverse. Work on \textit{inversion} in computer vision \cite{mahendran2014UnderstandingDI,dosovitskiy2016inverting} has shown that it is possible to learn to recover the input image (with some loss) given the logits of the final layer. Preliminary work has explored this question for text \cite{song2020informationleakage}, but only been able to recover an approximate bag of words given embeddings from shallow networks.

In this work, we target full reconstruction of input text from its embedding. If text is recoverable, there is a threat to privacy: a malicious user with access to a vector database, and text-embedding pairs from the model used to produce the data, could learn a function that reproduces text from embeddings.

We frame this problem of recovering textual embeddings as a controlled generation problem, where we seek to generate text such that the text is as close as possible to a given embedding. Our method, \textit{\MethodName}, uses the difference between a hypothesis embedding and a ground-truth embedding to make discrete updates to the text hypothesis. 

When we embed web documents using a state-of-the-art black-box encoder, our method can recover 32-token inputs with a near-perfect BLEU score of $97.3$, and can recover $92\%$ of the examples exactly. We then evaluate on embeddings generated from a variety of common retrieval corpuses from the BEIR benchmark. Even though these texts were not seen during training, our method is able to perfectly recover the inputs for a number of datapoints across a variety of domains. We evaluate on embeddings of clinical notes from MIMIC and are able to recover $89\%$ of full names from embedded notes. These results imply that text embeddings present the same threats to privacy as the text from which they are computed, and embeddings should be treated with the same precautions as raw data.

\section{Overview: Embedding Inversion}

Text embedding models learn to map text sequences to embedding vectors. Embedding vectors are useful because they encode some notion of semantic similarity: inputs that are similar in meaning should have embeddings that are close in vector space \cite{mikolov2013efficient}. Embeddings are commonly used for many tasks such as search, clustering, and classification \cite{aggarwal2012, neelakantan2022textopenai, muennighoff2023mteb}.

Given a text sequence of tokens $x \in \mathbb{V}^{n}$, a text encoder $\phi : \mathbb{V}^{n} \rightarrow \mathbb{R}^d$ maps $x$ to a fixed-length embedding vector $e \in \mathbb{R}^d$.

Now consider the problem of inverting textual embeddings: given some unknown encoder $\phi$, we seek to recover the text $x$ given its embedding $e = \phi(x)$. Text embedding models are typically trained to encourage similarity between related inputs \cite{karpukhin2020dpr}. Thus, we can write the problem as recovering text that has a maximally similar embedding to the ground-truth. We can formalize the search for text $\hat{x}$ with embedding $e$ under encoder $\phi$ as optimization:

\begin{equation}
\label{eq:prob}
\hat{x} = \arg\max_{x} \text{cos}(\phi(x), e)
\end{equation}

\paragraph{Assumptions of our threat model.} In a practical sense, we consider the scenario where an attacker wants to invert a single embedding produced from a black-box embedder $\phi$. We assume that the attacker has access to $\phi$: given hypothesis text $\hat{x}$, the attacker can query the model for $\phi(\hat{x})$ and compute $\text{cos}( \phi(\hat{x}), e)$. When this term is 1 exactly, the attacker can be sure that $\hat{x}$ was the original text, i.e. collisions are rare and can be ignored.

\section{Method: Vec2Text}
\label{sec:method}

\subsection{Base Model: Learning to Invert $\phi$}
\label{sec:simple}

Enumerating all possible sequences to compute \Cref{eq:prob} is computationally infeasible. One way to avoid this computational constraint is by learning a distribution of texts given embeddings. Given a dataset of texts $\mathcal{D} = \{x_1, \ldots \}$, we learn to invert encoder $\phi$ by learning a distribution of texts given embeddings, $p(x \mid e; \theta)$, by learning $\theta$ via maximum likelihood:
\[
\theta = \arg\,\max_{\hat{\theta}} \mathbb{E}_{x \sim \mathcal{D}} [ p(x \mid \phi(x); \hat{\theta}) ]
\]

\noindent We drop the $\theta$ hereon for simplicity of notation. In practice, this process involves training a conditional language model to reconstruct unknown text $x$ given its embedding $e = \phi(x)$. We can view this learning problem as amortizing the combinatorial optimization~(\Cref{eq:prob}) into the weights of a neural network. 
Directly learning to generate satisfactory text in this manner is well-known in the literature to be a difficult problem.

% Given a dataset of text samples $\mathcal{D}$, we make queries to $\phi$ to learn its inverse, i.e. $p(x \mid e)$ where $e = \phi(x)$. Parameterizing $\theta$ with a neural network and learning directly to produce text from embeddings is one possible approach. 

\subsection{Controlling Generation for Inversion}
\label{sec:congen}

% We nonetheless consider directly generating text from the embedding as a baseline in \Cref{sec:results}.
To improve upon this model, we propose \MethodName{} shown in Figure~\ref{fig:00_main}. This approach takes inspiration from methods for Controlled Generation, the task of generating text that satisfies a known condition~\cite{hu2018controlled, john2018disentangled, yang2021fudge}. This task is similar to inversion in that there is a observable function $\phi$ that determines the level of control. 
However, it differs in that approaches to controlled generation \cite{dathathri2020pplm,li2022diffusionlm} generally require differentiating through $\phi$ to improve the score of some intermediate representation. Textual inversion differs in that we can only make queries to $\phi$, and cannot compute its gradients. 

%  Note: Volo recommended to separate this section into Model, Learning, and Inference.

% Although we cannot compute $\phi$'s gradients, we can take inspiration from gradient-based methods for controlled text generation to develop a black-box approach. 

\paragraph{Model.} The method guesses an initial hypothesis and iteratively refines this hypothesis by re-embedding and correcting the hypothesis to bring its embedding closer to $e$. Note that this model requires computing a new embedding  $\hat{e}^{(t)}= \phi(x^{(t)})$ in order to generate each new correction $x^{(t+1)}$. We define our model recursively by marginalizing over intermediate hypotheses:

\begin{align*}
    p(x^{(t+1)} \mid e) &= \sum_{x^{(t)}}  p(x^{(t)} \mid e) p(x^{(t+1)} \mid e, x^{(t)}, \hat{e}^{(t)})\\
    \hat{e}^{(t)} &= \phi(x^{(t)})
\end{align*}

\noindent with a base case of the simple learned inversion:

\[
p(x^{(0)} \mid e) = p(x^{(0)} \mid e, \varnothing, \phi(\varnothing))
\]

\noindent %given ground-truth embedding $e$ and encoder $\phi$. 
Here, $x^{(0)}$ represents the initial hypothesis generation, $x^{(1)}$ the correction of $x^{(0)}$, and so on. We train this model by first generating hypotheses $x^{(0)}$ from the model in Section~\ref{sec:simple}, computing $\hat{e}^{(0)}$,  and then training a model on this generated data. 

% \paragraph{Related work in `iterative refinement'.} 
This method relates to other recent work generating text through iterative editing \cite{lee2018deterministic, ghazvininejad2019maskpredict}. Especially relevant is \citet{welleck2022generating}, which proposes to train a text-to-text `self-correction' module to improve language model generations with feedback. % Although we also train a model that corrects text, our model learns to map a `diff' in text embedding space to a discrete text correction, and can be recursively applied many times with continual improvement.

\paragraph{Parameterization.} The backbone of our model, $p(x^{(t+1)} \mid e, x^{(t)}, \hat{e}^{(t)})$, is parameterized as a standard encoder-decoder transformer \cite{vaswani2017attention,raffel2020t5} conditioned on the previous output. 

One challenge is the need to input conditioning embeddings $e$ and $\hat{e}^{(t)}$ into a transformer encoder, which requires a sequence of embeddings as input with some dimension $d_{\text{enc}}$ not necessarily equal to the dimension $d$ of $\phi$'s embeddings. 
Similar to \citet{mokady2021clipcap}, we use small MLP to project a single embedding vector to a larger size, and reshape to give it a sequence length as input to the encoder. For embedding $e \in \mathbb{R}^d$:

\[
\text{EmbToSeq}(e) = W_2 \  \sigma(W_1\   e)
\]

\noindent where $W_1 \in \mathbb{R}^{d \times d}$ and $W_2 \in \mathbb{R}^{(s d_{\text{enc}}) \times d}$ for some nonlinear activation function $\sigma$ and predetermined encoder ``length'' $s$. We use a separate MLP to project three vectors: the ground-truth embedding $e$, the hypothesis embedding $\hat{e}^{(t)}$, and the difference between these vectors $e - \hat{e}$. Given the word embeddings of the hypothesis $x^{(t)}$ are $\{w_1 ... w_n\}$, the input (length $3s + n$) to the encoder is as follows:
\begin{align*}
\text{concat}(
&\text{EmbToSeq}(e), \\
&\text{EmbToSeq}(\hat{e}^{(t)}), \\
& \text{EmbToSeq}(e - \hat{e}^{(t)}),  (w_1 ... w_n)
)
\end{align*}
We feed the concatenated input to the encoder and train the full encoder-decoder model using standard language modeling loss.

\paragraph{Inference.} In practice we cannot tractably sum out intermediate generations $x^{(t)}$, so we approximate this summation via beam search.
 We perform inference from our model greedily at the token level but implement beam search at the sequence-level $x^{(t)}$. At each step of correction, we consider some number $b$ of possible corrections as the next step. For each possible correction, we decode the top $b$ possible continuations, and then take the top $b$ unique continuations out of $b \cdot b$ potential continuations by measuring their distance in embedding space to the ground-truth embedding $e$.

\section{Experimental Setup}
\label{sec:experimental-setup}

\input{tables/01_results}
\input{tables/02_beir}

\paragraph{Embeddings.} \MethodName{} is trained to invert two state-of-the-art embedding models: GTR-base \cite{ni2021gtr}, a T5-based pre-trained transformer for text retrieval, and \texttt{text-embeddings-ada-002} available via the OpenAI API. Both model families are among the highest-performing embedders on the MTEB text embeddings benchmark \cite{muennighoff2023mteb}.

\paragraph{Datasets.} We train our GTR model on $5M$ passages from Wikipedia articles selected from the Natural Questions corpus \cite{kwiatkowski2019naturalquestions} truncated to $32$ tokens. We train our two OpenAI models \cite{bajaj2018msmarco}, both on versions of the MSMARCO corpus with maximum $32$ or $128$ tokens per example \footnote{By 2023 pricing of $\$0.0001$ per $1000$ tokens, embedding 5 million documents of 70 tokens each costs $\$35$.}. For evaluation, we consider the evaluation datasets from Natural Questions and MSMarco, as well as two out-of-domain settings: the MIMIC-III database of clinical notes \cite{johnson2016mimiciii} in addition to the variety of datasets available from the BEIR benchmark \cite{thakur2021beir}.

\paragraph{Baseline.} As a baseline, we train the base model $p(x^{(0)} \mid e)$ to recover text with no correction steps. We also evaluate the bag of words model from \citet{song2020informationleakage}. To balance for the increased number of queries allotted to the correction models, we also consider taking the top-N predictions made from the unconditional model via beam search and nucleus sampling $(p = 0.9)$ and reranking via cosine similarity.

\paragraph{Metrics.} We use two types of metrics to measure the progress and the accuracy of reconstructed text. First we consider our main goal of text reconstruction. To measure this we use word-match metrics including: BLEU score \cite{papineni2002bleu}, a measure of n-gram similarities between the true and reconstructed text; Token F1, the multi-class F1 score between the set of predicted tokens and the set of true tokens; Exact-match, the percentage of reconstructed outputs that perfectly match the ground-truth.
We also report the similarity on the internal inversion metric in terms of recovering the vector embedding in latent space. We use cosine similarity between the true embedding and the embedding of reconstructed text according to $\phi$.

\paragraph{Models and Inference.} We initialize our models from the T5-base checkpoint \cite{raffel2020t5}. Including the projection head, each model has approximately 235M parameters. We set the projection sequence length $s = 16$ for all experiments, as preliminary experiments show diminishing returns by increasing this number further. We perform inference on all models using greedy token-level decoding. When running multiple steps of sequence-level beam search, we only take a new generation if it is closer than the previous step in cosine similarity to the ground-truth embedding.

We use unconditional models to seed the initial hypothesis for our iterative models. We examine the effect of using a different initial hypothesis in \Cref{sec:analysis}.

We use the Adam optimizer and learning rate of $2*10^{-4}$ with warmup and linear decay. We train models for $100$ epochs. We use batch size of 128 and train all models on 4 NVIDIA A6000 GPUs. Under these conditions, training our slowest model takes about two days. 

\section{Results}
\label{sec:results}

\subsection{Reconstruction: In-Domain}
\Cref{table1:in_domain} contains in-domain results. Our method outperforms the baselines on all metrics. More rounds is monotonically helpful, although we see diminishing returns – we are able to recover $77\%$ of BLEU score in just 5 rounds of correction, although running for $50$ rounds indeed achieves a higher reconstruction performance. We find that running sequence-level beam search (sbeam) over the iterative reconstruction is particularly helpful for finding exact matches of reconstructions, increasing the exact match score by $2$ to $6$ times across the three settings. In a relative sense, the model has more trouble exactly recovering longer texts, but still is able to get many of the words.

% Clearly, our model learns to make discrete updates to sequence $x^{(t)}$ based on its direction and distance in embedding space from $e$ and move it closer. 

\subsection{Reconstruction: Out-of-Domain}

We evaluate our model on $15$ datasets from the BEIR benchmark and display results in \Cref{table2:beir}. Quora, the shortest dataset in BEIR, is the easiest to reconstruct, and our model is able to exactly recover $66\%$ of examples. Our model adapts well to different-length inputs, generally producing reconstructions with average length error of fewer than $3$ tokens. In general, reconstruction accuracy inversely correlates with example length (discussed more in \Cref{sec:analysis}). On all datasets, we are able to recover sequences with Token F1 of at least $41$ and cosine similarity to the true embedding of at least $0.95$. 

\subsection{Case study: MIMIC}
\label{results:mimic}

As a specific threat domain, we consider MIMIC-III clinical notes \cite{johnson2016mimiciii}. Because the original release of MIMIC is completely deidentified, we instead use the ``pseudo re-identified'' version from \citet{lehman2021does} where fake names have been inserted in the place of the deidentified ones. 

Each note is truncated to 32 tokens and the notes are filtered so that they each contain at least one name. We measure the typical statistics of our method as well as three new ones: the percentage of first names, last names, and complete names that are recovered. Results are shown in \Cref{table3:mimic}. \MethodName\ is able to recover $94\%$ of first names, $95\%$ of last names, and $89\%$ of full names (first, last format) while recovering $26\%$ of the documents exactly. 
% We analyze \MethodName's ability to recover specific entities from MIMIC embeddings in \Cref{sec:analysis}.

For the recovered clinical notes from \Cref{results:mimic}, we extract entities from each true and recovered note using a clinical entity extractor \cite{raza2022entity}. We plot the recovery percentage in \ref{table3:mimic} (bottom) with the average entity recovery shown as a dashed line. Our model is most accurate at reconstructing entities of the type ``Clinical Event'', which include generic medical words like `arrived', `progress', and `transferred'. Our model is least accurate in the ``Detailed Description'' category, which includes specific medical terminology like `posterior' and `hypoxic', as well as multi-word events like `invasive ventilation - stop 4:00 pm'. 

Although we are able to recover $26\%$ of $32$-token notes exactly, the notes that were not exactly recovered are semantically close to the original. Our model generally matches the syntax of notes, even when some entities are slightly garbled; for example, given the following sentence from a doctor's note ``Rhona Arntson npn/- \# resp: infant remains orally intubated on imv / r fi'' our model predicts ``Rhona Arpson nrft:\# infant remains intubated orally on resp. imv. m/n fi''.

\input{tables/03_mimic}

\section{Defending against inversion attacks}
Is it easy for users of text embedding models protect their embeddings from inversion attacks? We consider a basic defense scenario as a sanity check. To implement our defense,  the user addes a level of Gaussian noise directly to each embedding with the goal of effectively defending against inversion attacks while preserving utility in the nearest-neightbor retrieval setting. We analyze the trade-off between retrieval performance and reconstruction accuracy under varying levels of noise.

Formally, we define a new embedding model as:
\[
\phi_{\text{noisy}}(x) = \phi(x) + \lambda \cdot \epsilon, \epsilon \sim N(0, 1)
\]
\noindent where $\lambda$ is a hyperparameter controlling the amount of noise injected. 

We simulate this scenario with $\phi$ as GTR-base using our self-corrective model with $10$ steps of correction, given the noisy embedder $\phi_{\text{noisy}}$. To measure retrieval performance, we take the mean NDCG@10 (a metric of retrieval performance; higher is better) across $15$ different retrieval tasks from the BEIR benchmark, evaluated across varying levels of noise. 

\begin{figure}[t]
    \centering
    % \vspace{-25pt}
    \includegraphics[width=0.9\columnwidth]{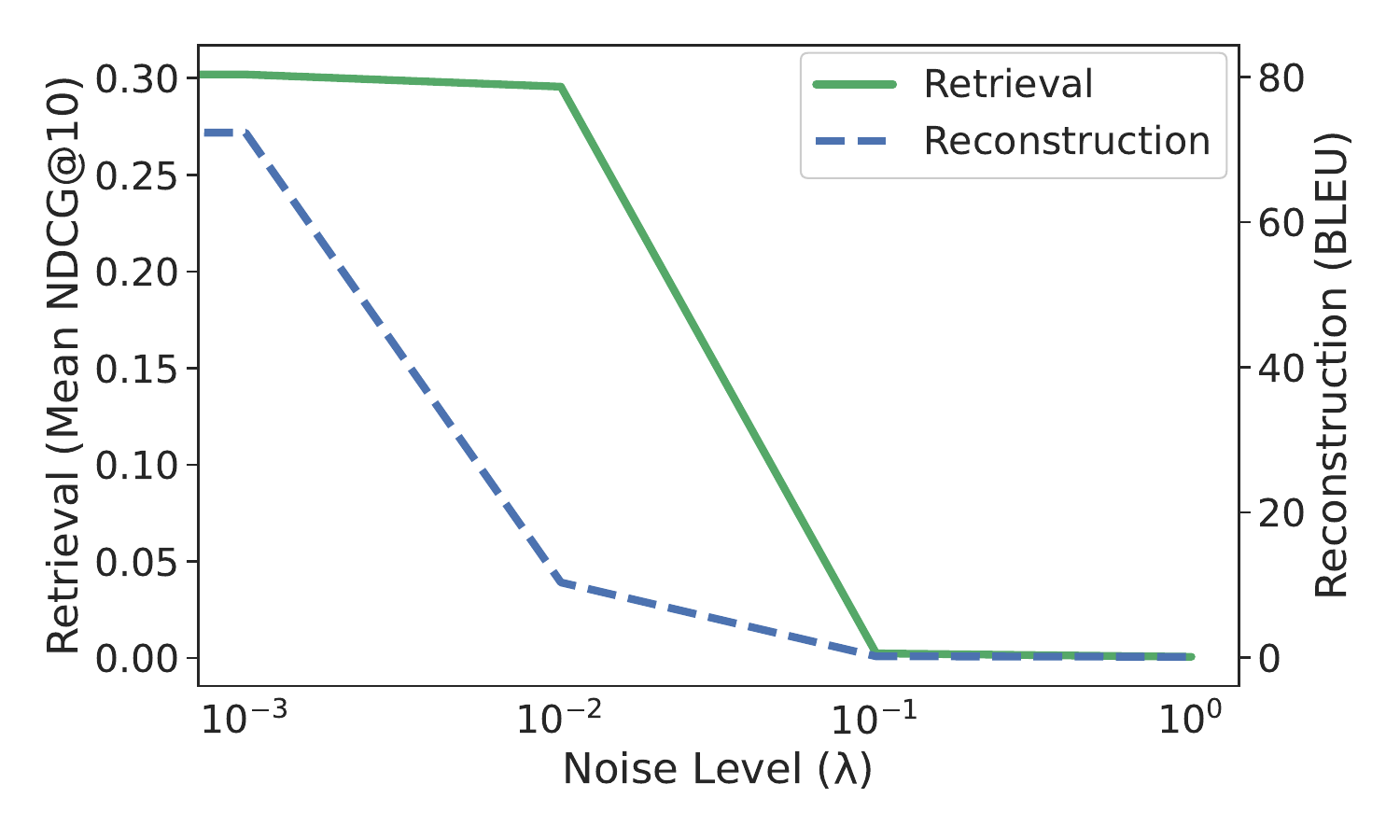}
    \caption{Retrieval performance and reconstruction accuracy across varying levels of noise injection.}
    \label{fig:04_retrieval_under_noise}
\end{figure}

We graph the average retrieval performance in \Cref{fig:04_retrieval_under_noise} (see \ref{app:defense-results} for complete tables of results). 
At a noise level of $\lambda = 10^{-1}$, we see retrieval performance is preserved, while BLEU score drops by $10\%$. At a noise level of $0.01$, retrieval performance is barely degraded ($2\%$) while reconstruction performance plummets to $13\%$ of the original BLEU. Adding any additional noise severely impacts both retrieval performance and reconstruction accuracy. These results indicate that adding a small amount of Gaussian noise may be a straightforward way to defend against naive inversion attacks, although it is possible that training with noise could in theory help \MethodName\ recover more accurately from $\phi_{noisy}$. Note that low reconstruction BLEU score is not necessarily indicative that coarser inferences, such as clinical area or treatment regimen, cannot be made from embeddings.

\section{Analysis}
\label{sec:analysis}

% Privacy experiment ideas
% Recovering JSON?
% Recovering Names/Dates/Other entities?
% ???

% \paragraph{Which entities are easiest to recover?}

\begin{figure}[t]
    \centering
    % \vspace{-25pt}
    \includegraphics[width=0.9\columnwidth]{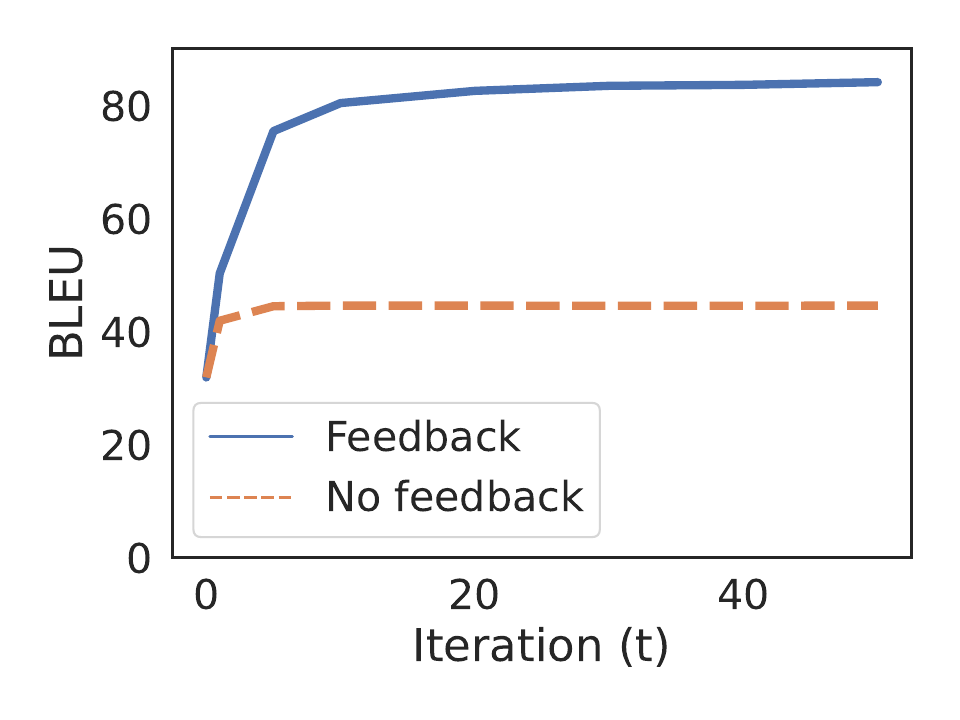}
    \caption{
    Recovery performance across multiple rounds of self-correction comparing models with access to $\phi$ vs text-only (32 tokens per sequence).
    }
    \label{fig:01_multiround}
\end{figure}

\paragraph{How much does the model rely on feedback from $\phi$?} \Cref{fig:01_multiround} shows an ablation study of the importance of feedback, i.e. performing corrections with and without embedding the most recent hypothesis. The model trained with feedback (i.e. additional conditioning on $\phi(x^{(t)})$ is able to make a more accurate first correction and gets better BLEU score with more rounds. The model trained with no feedback can still edit the text but does not receive more information about the geometry of the embedding space, and quickly plateaus. The most startling comparison is in terms of the number of exact matches: after 50 rounds of greedy self-correction, our model with feedback gets $52.0\%$ of examples correct (after only $1.5\%$ initially); the model trained without feedback only perfectly matches $4.2\%$ of examples after $50$ rounds.

% \paragraph{Why does our model work recursively?} 

\begin{figure}[t]
    \centering
    % \vspace{-25pt}
    \includegraphics[width=0.9\columnwidth]{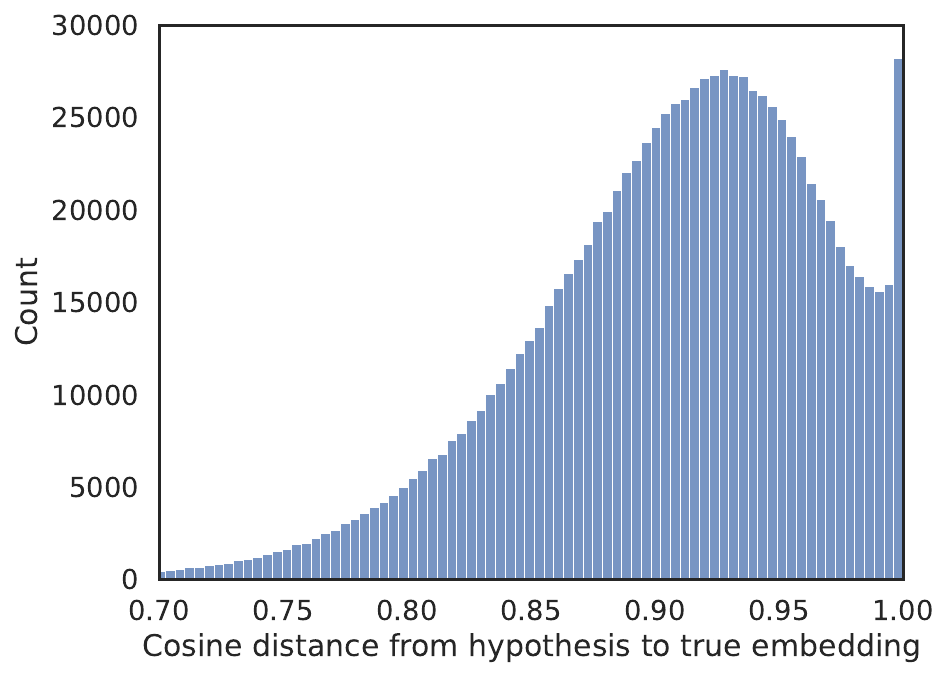}
    \caption{
    Distribution of $\cos(e, \phi(x^{(0)}))$ over training data. The mean training output from the GTR base model has a cosine similarity of $0.924$ with the true embedding.
    }
    \label{fig:02_hypothesis_closeness}
\end{figure}

During training, the model only learns to correct a single hypothesis to the ground-truth sample. Given new text at test time, our model is able to correct the same text multiple times, ``pushing'' the text from $0.9$ embedding similarity to $1.0$. We plot the closeness of the first hypothesis to the ground-truth in the training data for the length-32 model  in \Cref{fig:02_hypothesis_closeness}. We see that during training the model learns to correct hypotheses across a wide range of closenesses, implying that corrections should not go `out-of-distribution' as they approach the ground-truth.

\input{tables/05_qualitative_example}

\begin{figure}[t]
    \centering
    % \vspace{-25pt}
    \includegraphics[width=0.9\columnwidth]{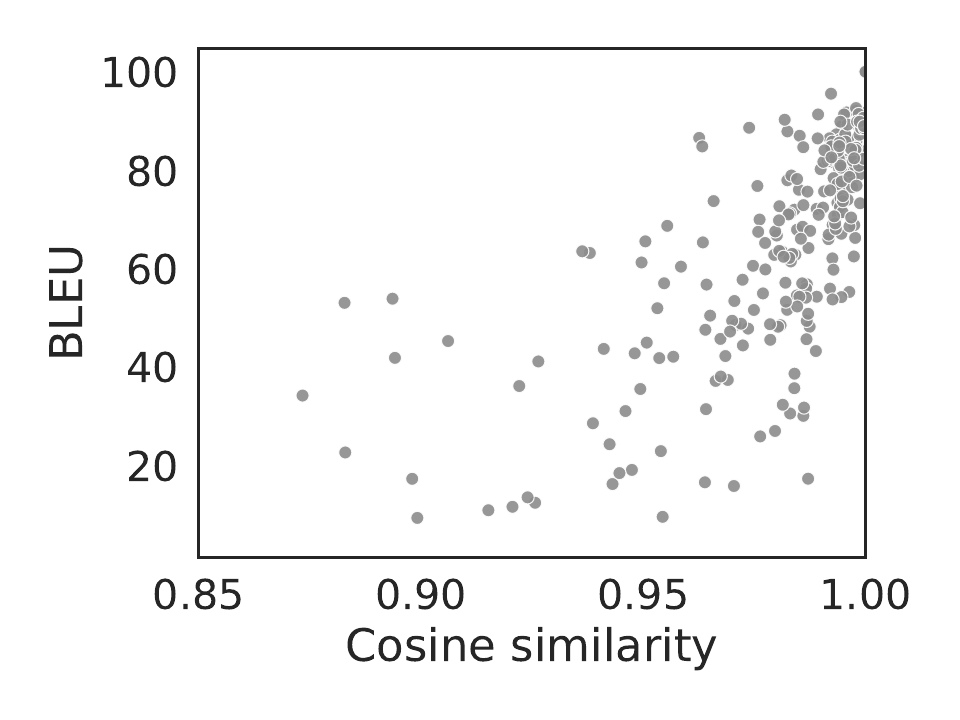}
    \caption{Cosine similarity vs BLEU score on $1000$ reconstructed embeddings from Natural Questions text.}
    \label{fig:05_cos_sim_vs_bleu}
\end{figure}

\paragraph{How informative are embeddings for textual recovery? }
We graph BLEU score vs. cosine similarity from a selection of of reconstructed text inputs in \Cref{fig:05_cos_sim_vs_bleu}. We observe a strong correlation between the two metrics. Notably, there are very few generated samples with high cosine similarity but low BLEU score. This implies that better following embedding geometry will further improves  systems.
% mention Goodhart's law? like in RL4LMs intro?
Theoretically some embeddings might be impossible to recover. Prior work \cite{song2020adversarial, morris2020secondorder} has shown that two different sequences can `collide' in text embedding space, having similar embeddings even without any word overlap. 
However, our experiments found no evidence that collisions are a problem; they either do not exist or our model learns during training to avoid outputting them. Improved systems should be able to recover longer text.

\paragraph{Does having a strong base model matter?}
We ablate the impact of initialization by evaluating our $32$-token Wikipedia model at different initializations of $x^{(0)}$, as shown in \Cref{table4:ablation}. After running for $20$ steps of correction, our model is able to recover from an unhelpful initialization, even when the initialization is a random sequence of tokens. This suggests that the model is able to ignore bad hypotheses and focus on the true embedding when the hypothesis is not helpful.

\input{tables/04_init_ablation}

\section{Related work}

\paragraph{Inverting deep embeddings.} The task of inverting textual embeddings is closely related to research on inverting deep visual representations in computer vision \cite{mahendran2014UnderstandingDI,dosovitskiy2016inverting,teterwak2021understanding,bordes2021highfissl}, which show that a high amount of visual detail remains in the logit vector of an image classifier, and attempt to reconstruct input images from this vector. There is also a line of work reverse-engineering the content of certain text embeddings: \citet{ram2023token} analyze the contents of text embeddings by projecting embeddings into the model's vocabulary space to produce a distribution of relevant tokens. \citet{adolphs2022decoding} train a single-step query decoder to predict the text of queries from their embeddings and use the decoder to produce more data to train a new retrieval model. We focus directly on text reconstruction and its implications for privacy, and propose an iterative method that works for paragraph-length documents, not just sentence-length queries.

\paragraph{Privacy leakage from embeddings.} Research has raised the question of information leakage from dense embeddings. In vision, Vec2Face \cite{duong2020vec2face} shows that faces can be reconstructed from their deep embeddings. Similar questions have been asked about text data:
\citet{lehman2021does} attempt to recover sensitive information such as names from representations obtained from a model pre-trained on clinical notes, but fail to recover exact text. \citet{kim2022towardhomomorphic} propose a privacy-preserving similarity mechanism for text embeddings and consider a shallow bag-of-words inversion model. \citet{abdalla2020wordembeddingprivacy} analyze the privacy leaks in training word embeddings on medical data and are able to recover full names in the training data from learned word embeddings. \citet{dziedzic2023sentence} note that \textit{stealing} sentence encoders by distilling through API queries works well and is difficult for API providers to prevent. \citet{song2020informationleakage} considered the problem of recovering text sequences from embeddings, but only attempted to recover bags of words from the embeddings of a shallow encoder model. \citet{li2023sentence} investigate the privacy leakage of embeddings by training a decoder with a text embedding as the first embedding fed to the decoder. Compared to these works, we consider the significantly more involved problem of developing a method to recover the full ordered text sequence from more realistic state-of-the-art text retrieval models.

\paragraph{Gradient leakage.} 
There are parallels between the use of vector databases to store embeddings and the practice of federated learning, where users share gradients with one another in order to jointly train a model. Our work on analyzing the privacy leakage of text embeddings is analogous to research on \textit{gradient leakage}, which has shown that certain input data can be reverse-engineered from the model's gradients during training \cite{melis2018exploiting,zhu2019deepgradientleakage,zhao2020idlg,geiping2020inverting}. 
\citet{zhu2019deepgradientleakage} even shows that they can recover text inputs of a masked language model by backpropagating to the input layer to match the gradient. However, such techniques do not apply to textual inversion: the gradient of the model is relatively high-resolution; we consider the more difficult problem of recovering the full input text given only a single dense embedding vector.

\paragraph{Text autoencoders.} Past research has explored natural language processing learning models that map vectors to sentences \cite{bowman2016generating}. These include some retrieval models that are trained with a shallow decoder to reconstruct the text or bag-of-words from the encoder-outputted embedding \cite{xiao2022retromae,shen2023lexmae,wang2023simlm}. Unlike these, we invert embeddings from a frozen, pre-trained encoder.

\section{Conclusion}

We propose \MethodName, a multi-step method that iteratively corrects and re-embeds text based on a fixed point in latent space. Our approach can recover $92\%$  of $32$-token text inputs from their embeddings exactly, demonstrating that text embeddings reveal much of the original text. The model also demonstrates the ability to extract critical clinical information from clinical notes, highlighting its implications for data privacy in sensitive domains like medicine. 

Our findings indicate a sort of equivalence between embeddings and raw data, in that both leak similar amounts of sensitive information. This equivalence puts a heavy burden on anonymization requirements for dense embeddings: embeddings should be treated as highly sensitive private data and protected, technically and perhaps legally, in the same way as one would protect raw text.

\section{Limitations}
% Note: limitations doesn't count towards the page limit:
% https://2023.emnlp.org/calls/main_conference_papers/#mandatory-discussion-of-limitations

\paragraph{Adaptive attacks and defenses.} We consider the setting where an adversary applies noise to newly generated embeddings, but the reconstruction modules were trained from un-noised embeddings. Future work might consider reconstruction attacks or defenses that are adaptive to the type of attack or defense being used.

\paragraph{Search thoroughness.} Our search is limited; in this work we do not test beyond searching for $50$ rounds or with a sequence beam width higher than $8$. However,\ \MethodName\ gets monotonically better with more searching. Future work could find even more exact matches by searching for more rounds with a higher beam width, or by implementing more sophisticated search algorithms on top of our corrective module.

\paragraph{Scalability to long text.} Our method is shown to recover most sequences exactly up to $32$ tokens and some information up to $128$ tokens, but we have not investigated the limits of inversion beyond embeddings of this length. Popular embedding models support embedding text content on the order of thousands of tokens, and embedding longer texts is common practice \cite{thakur2021beir}. Future work might explore the potential and difficulties of inverting embeddings of these longer texts.

\paragraph{Access to embedding model.} Our threat model assumes that an adversary has black-box access to the model used to generate the embeddings in the compromised database. In the real world, this is realistic because practitioners so often rely on the same few large models. However, \MethodName\ requires making a query to the black-box embedding model for each step of refinement. Future work might explore training an imitation embedding model which could be queried at inference time to save queries to the true embedder.

% Acknowledgements: Helpful discussions with Piotr, Simran, Justin Chiu

\bibliography{anthology,custom,vector_databases}
\bibliographystyle{acl_natbib}

\appendix

\section{Appendix}
\label{sec:appendix}

\subsection{Additional analysis}

% srush comment on this: too niche, either combine with something else or removet
% \paragraph{How does BLEU correlate with model likelihood?}
% What is the correlation between model likelihood and BLEU score? Are we making search errors? How does this change during correction? {\color{red} Add graph of BLEU score vs. likelihood for a number of sampled sequences.}

\begin{figure}[t]
    \centering
    % \vspace{-25pt}
    \includegraphics[width=0.9\columnwidth]{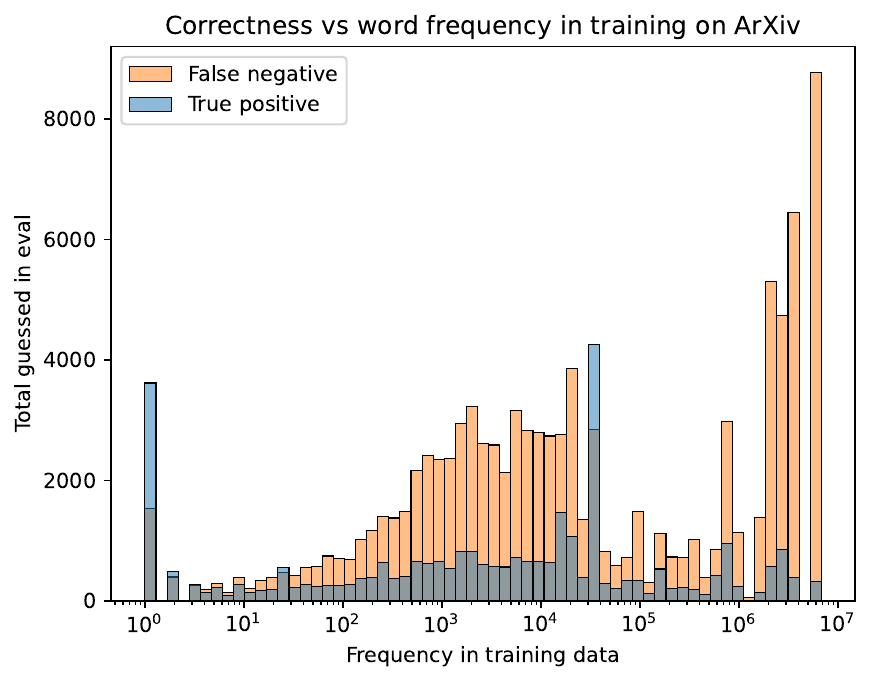}
    \caption{Correctness on evaluation samples from ArXiv data.}
    \label{fig:03_incorrect_by_frequency}
\end{figure}

\paragraph{How does word frequency affect model correctness?} \cref{fig:03_incorrect_by_frequency} shows the number of correct predictions (orange) and incorrect predictions (blue) for ground-truth words, plotted across word frequency in the training data. Our model generally predicts words better that are more frequent in the training data, although it is still able to predict correctly a number of words that were not seen during training\footnote{We hypothesize this is because all test \textit{tokens} were present in the training data, and the model is able to reconstruct unseen words from seen tokens.}. Peaks between $10^4$ and $10^5$ come from the characters $($, $-$, and $)$, which appear frequently in the training data, but are still often guessed incorrectly in the reconstructions.

% \paragraph{How does sequence length affect recoverability?} 

% \Cref{fig:06_bleu_vs_num_tokens} plots the performance of our model at reconstructing text from OpenAI embeddings on datasets from the BEIR benchmark across the number of tokens in the dataset. We observe a negative correlation: as the number of tokens represented by the (fixed-size) embedding increases, the original sequence is more difficult to reconstruct. Quora, which is by far the shortest of the datasets, is the easiest to reconstruct.

% \begin{figure}[t]
%     \centering
%     % \vspace{-25pt}
%     \includegraphics[width=0.9\columnwidth]{figs/06_bleu_vs_num_tokens.pdf}
%     \caption{Reconstruction performance across the $19$ datasets of BEIR.}
%     \label{fig:06_bleu_vs_num_tokens}
% \end{figure}

\subsection{Full defense results}
\label{app:defense-results}
Results on each dataset from BEIR under varying levels of Gaussian noise are shown in \Cref{app:tab:defense_retrieval_results}. The model is GTR-base. Note that the inputs are limited to $32 tokens$, far shorter than the average length for some corpuses, which is why baseline ($\lambda = 0$) NDCG@10 numbers are lower than typically reported. We included the full results (visualized in \Cref{fig:04_retrieval_under_noise}) as \Cref{app:tab:defense_aggregate_results}.

\input{tables/app_01_defense_retrieval}
\input{tables/app_02_defense_agg}

\end{document}

%% file: tables/01_results.tex
\begin{table*}[t]
\centering
% \resizebox{\textwidth}{!}{
\begin{tabular}{@{}@{}llllrrrr}
\toprule
% \multicolumn{5}{l}{\textbf{wikipedia}} \\
& method & tokens & pred tokens & bleu & tf1 & exact & cos\\
\midrule
\multirow{9}{*}{{\rotatebox[origin=c]{90}{GTR}}}
\multirow{9}{*}{{\rotatebox[origin=c]{90}{Natural Questions}}}
& Bag-of-words \cite{song2020informationleakage} & 32 & 32 & 0.3 & 51 & 0.0 & 0.70  \\
& GPT-2 Decoder \cite{li2023sentence} & 32 & 32 & 1.0 & 47 & 0.0 & 0.76  \\
& Base [0 steps] & 32 & 32 & 31.9 & 67 & 0.0 & 0.91 \\
& \quad (+ beam search) & 32 & 32 & 34.5 & 67 & 1.0 & 0.92\\
& \quad (+ nucleus)  & 32 & 32 & 25.3 & 60 & 0.0 & 0.88 \\
& \MethodName\ [1 step] & 32 & 32 & 50.7 & 80 & 0.0 & 0.96 \\
& \quad [20 steps] & 32 & 32 & 83.9 & 96 & 40.2 & 0.99   \\
& \quad [50 steps] & 32 & 32 & 85.4 & 97 & 40.6 & 0.99  \\ 
& \quad [50 steps + sbeam] & 32 & 32 & \textbf{97.3} & \textbf{99} & \textbf{92.0} & \textbf{0.99}  \\ 
\midrule
\multirow{5}{*}{{\rotatebox[origin=c]{90}{{OpenAI}}}}
\multirow{5}{*}{{\rotatebox[origin=c]{90}{MSMARCO}}}
& Base [0 steps] & 31.8 & 31.8 & 26.2 & 61 & 0.0 & 0.94 \\
& \MethodName\ [1 step] & 31.8 & 31.9 & 44.1 & 77 & 5.2 & 0.96 \\
& \quad [20 steps] & 31.8 & 31.9 & 61.9 & 87 & 15.0 & 0.98  \\
& \quad [50 steps] & 31.8 & 31.9 & 62.3 & 87 & 14.8 & 0.98 \\
& \quad [50 steps + sbeam] & 31.8 & 31.8 & \textbf{83.4} & \textbf{96} & \textbf{60.9} & \textbf{0.99} \\
\midrule
\multirow{5}{*}{{\rotatebox[origin=c]{90}{OpenAI}}}
\multirow{5}{*}{{\rotatebox[origin=c]{90}{MSMARCO}}}
& Base [0 steps] & 80.9 & 84.2 & 17.0 & 54 & 0.6 & 0.95  \\
& \MethodName\ [1 step] & 80.9 & 81.6 & 29.9 & 68 & 1.4 & 0.97 \\
& \quad [20 steps] & 80.9 & 79.7 & 43.1 & 78 & 3.2 & 0.99 \\
& \quad [50 steps] & 80.9 & 80.5 & 44.4 & 78 & 3.4 & 0.99 \\
& \quad [50 steps + sbeam] & 80.9 & 80.6 & \textbf{55.0} & \textbf{84} & \textbf{8.0} & \textbf{0.99} \\
\bottomrule
\end{tabular}
% }
\caption{Reconstruction score on in-domain datasets. Top section of results come from models trained to reconstruct $32$ tokens of text from Wikpedia, embedded using GTR-base. Remaining results come from models trained to reconstruct up to $32$ or $128$ tokens from MSMARCO, embedded using OpenAI \texttt{text-embeddings-ada-002}.}
\label{table1:in_domain}
\end{table*}

%% file: tables/02_beir.tex
\begin{table}[t!]
    \centering
    \resizebox{\linewidth}{!}{
        \begin{tabular}{llrrr}
        \toprule

              dataset & tokens & method & bleu & token F1 \\
        \midrule
\multirow{2}{*}{quora} & \multirow{2}{*}{15.7} & Base & 36.2 & 73.8 \\
& & Vec2Text & 95.5 & 98.6 \\
\midrule
\multirow{2}{*}{signal1m} & \multirow{2}{*}{23.7} & Base & 13.2 & 49.5 \\
& & Vec2Text & 80.7 & 92.5 \\
\midrule
\multirow{2}{*}{msmarco} & \multirow{2}{*}{72.1} & Base & 15.5 & 54.1 \\
& & Vec2Text & 59.6 & 86.1 \\
\midrule
\multirow{2}{*}{climate-fever} & \multirow{2}{*}{73.4} & Base & 12.8 & 49.3 \\
& & Vec2Text & 44.9 & 82.6 \\
\midrule
\multirow{2}{*}{fever} & \multirow{2}{*}{73.4} & Base & 12.6 & 49.2 \\
& & Vec2Text & 45.1 & 82.7 \\
\midrule
\multirow{2}{*}{dbpedia-entity} & \multirow{2}{*}{91.3} & Base & 15.4 & 50.3 \\
& & Vec2Text & 48.0 & 77.9 \\
\midrule
\multirow{2}{*}{nq} & \multirow{2}{*}{94.7} & Base & 11.0 & 47.1 \\
& & Vec2Text & 32.7 & 72.7 \\
\midrule
\multirow{2}{*}{hotpotqa} & \multirow{2}{*}{94.8} & Base & 15.4 & 50.1 \\
& & Vec2Text & 46.6 & 78.7 \\
\midrule
\multirow{2}{*}{fiqa} & \multirow{2}{*}{103.8} & Base & 6.6 & 44.1 \\
& & Vec2Text & 21.5 & 63.6 \\
\midrule
\multirow{2}{*}{webis-touche2020} & \multirow{2}{*}{105.2} & Base & 6.6 & 41.5 \\
& & Vec2Text & 19.6 & 69.7 \\
\midrule
\multirow{2}{*}{cqadupstack} & \multirow{2}{*}{106.4} & Base & 7.1 & 41.5 \\
& & Vec2Text & 23.3 & 64.3 \\
\midrule
\multirow{2}{*}{arguana} & \multirow{2}{*}{113.5} & Base & 6.8 & 44.1 \\
& & Vec2Text & 23.4 & 66.3 \\
\midrule
\multirow{2}{*}{scidocs} & \multirow{2}{*}{125.3} & Base & 5.9 & 38.5 \\
& & Vec2Text & 17.7 & 57.6 \\
\midrule
\multirow{2}{*}{trec-covid} & \multirow{2}{*}{125.4} & Base & 5.6 & 36.3 \\
& & Vec2Text & 19.3 & 58.6 \\
\midrule
\multirow{2}{*}{robust04} & \multirow{2}{*}{127.3} &Base & 4.9 & 34.4 \\
 & & Vec2Text & 15.5 & 54.5 \\
 \midrule
\multirow{2}{*}{bioasq} & \multirow{2}{*}{127.4} & Base & 5.3 & 35.7 \\
& & Vec2Text & 22.8 & 59.5 \\
\midrule
\multirow{2}{*}{scifact} & \multirow{2}{*}{127.4} & Base & 4.9 & 35.2 \\
& & Vec2Text & 16.6 & 56.6 \\
\midrule
\multirow{2}{*}{nfcorpus} & \multirow{2}{*}{127.7} & Base & 6.2 & 39.6 \\
& & Vec2Text & 25.8 & 64.8 \\
\midrule
\multirow{2}{*}{trec-news} & \multirow{2}{*}{128.0} & Base & 4.9 & 34.8 \\
& & Vec2Text & 14.5 & 51.5 \\
\bottomrule
    
    % dataset & tokens & ptokens & tf1 & cos \\
    %     \midrule
    %     quora & 15.5 & 18.6 & 93.5 & 99  \\
    %     signal1m & 24.5 & 27.2 & 73.3 & 97  \\
    %     msmarco & 72.1 & 73.8 & 71.5 & 98\\
    %     fever & 73.4 & 71.8 & 70.8 & 98 \\
    %     dbpedia-entity & 93.1 & 92.6 & 65.0 & 98\\
    %     hotpotqa & 94.3 & 94.2 & 64.3 & 98  \\
    %     nq & 95.5 & 93.9 & 60.6 & 97\\
    %     webis-touche2020 & 105.2 & 105.0 & 56.2 & 95 \\
    %     cqadupstack & 106.3 & 106.1 & 51.6 & 94 \\
    %     scidocs & 124.5 & 122.8 & 46.1 & 96 \\
    %     trec-covid & 125.2 & 123.5 & 47.0 & 96 \\
    %     scifact & 127.2 & 125.7 & 44.4 & 96 \\
    %     nfcorpus & 127.3 & 125.8 & 50.6 & 97 \\
    %     bioasq & 127.4 & 125.4 & 45.3 & 96 \\
    %     trec-news & 128.0 & 124.8 & 41.2 & 95 \\
        % \bottomrule
        \end{tabular}
    }
    \caption{Out-of-domain reconstruction performance measured on datasets from the BEIR benchmark. We sort datasets in order of average length in order to emphasize the effect of sequence length on task difficulty.}
\label{table2:beir}
\end{table}

%% file: tables/03_mimic.tex
\begin{table}[t]
\centering
\resizebox{\columnwidth}{!}{
\begin{tabular}{@{}llllrrrrrr}
    \toprule
    % \multicolumn{5}{l}{\textbf{wikipedia}} \\
    & method & first & last & full & bleu & tf1 & exact & cos\\
    \midrule
    & Base  & 40.0 & 27.8 & 10.8 & 4.9 & 33.1 & 0. & 0.78 \\
    & \MethodName & 94.2 & 95.3 & 89.2 & 55.6 & 80.8 & 26.0 & 0.98 \\
    \bottomrule
\end{tabular}
}

% \begin{figure}[t]
    \centering
    % \vspace{-25pt}
    \includegraphics[width=\columnwidth]{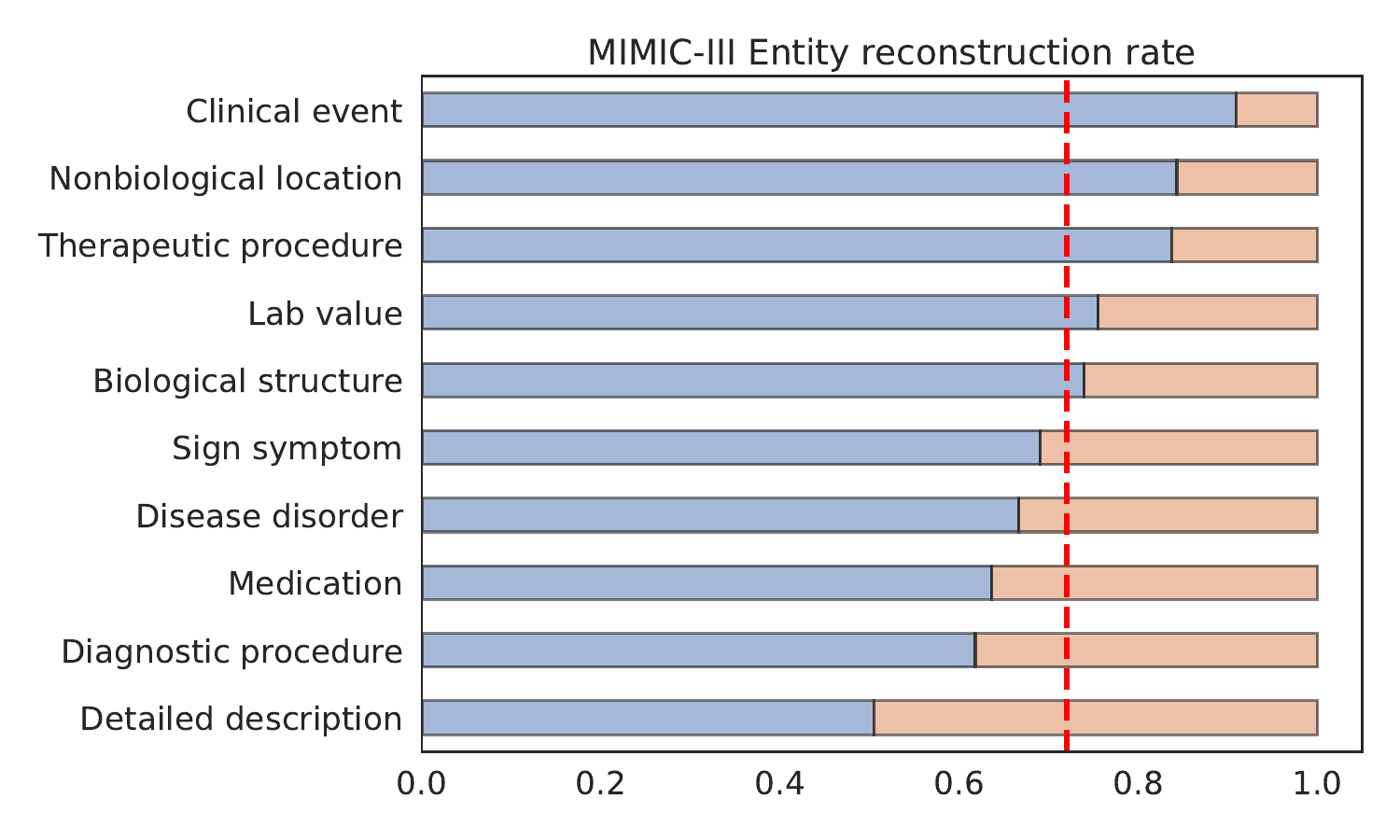}
    % \caption{
    % Entity recovery performance on $32$-token clinical notes from MIMIC-III.
    % }
% \end{figure}
\caption{Performance of our method on reconstructing GTR-embedded clinical notes from MIMIC III \cite{johnson2016mimiciii}.}
\label{table3:mimic}
\label{fig:07_mimic_entities}

\end{table}

%% file: tables/05_qualitative_example.tex
\begin{table*}[t]
\small
\centering
    % \resizebox{\textwidth}{!}{
        \begin{tabular}{@{}lll@{}}
        \toprule
        Input & Nabo Gass (25 August, 1954 in Ebingen, Germany) is a German painter and glass artist. & \\ \midrule
        Round 1 (0.85): & Nabo Gass (11 August 1974 in Erlangen, Germany) is an artist. & {\color{red}\ding{55}} \\
        Round 2 (0.99): & Nabo Gass (b. 18 August 1954 in Egeland, Germany) is a German painter and glass artist. & {\color{red}\ding{55}} \\
        Round 3 (0.99): & Nabo Gass (25 August 1954 in Ebingen, Germany) is a German painter and glass artist. & {\color{red}\ding{55}} \\
        Round 4 (1.00): & Nabo Gass (25 August, 1954 in Ebingen, Germany) is a German painter and glass artist. & {\color{green}\ding{51}} \\ \bottomrule
        \end{tabular}
        \caption{Example of our corrective model working in multiple rounds. Left column shows the correction number, from Round 1 (initial hypothesis) to Round 4 (correct guess). The number in parenthesis is the cosine similarity between the guess's embedding and the embedding of the ground-truth sequence (first row).}
    % }
\end{table*}

%% file: tables/04_init_ablation.tex
\begin{table}[t!]
% \small
\centering
\resizebox{\columnwidth}{!}{
    \begin{tabular}{llll}
    \toprule
    Initialization & token f1 & cos & exact \\
    \midrule
    Random tokens & 0.95 & 0.99 & 50.0 \\
    "the " * 32 & 0.95 & 0.99 & 49.8 \\
    "there's no reverse on a motorcycle, \\as my friend found out quite\\ dramatically the other day" & 0.96 & 0.99 & 52.0 \\
    \midrule
    Base model $p(x^{(0)} \mid e)$ & 0.96 & 0.99 & 51.6 \\
    \bottomrule
    \end{tabular}
}
\label{table4:ablation}
\caption{Ablation: Reconstruction score on Wikipedia data ($32$ tokens) given various initializations. Our self-correction model is able to faithfully recover the original text with greater than $80$ BLEU score, even with a poor initialization. Models run for $20$ steps of correction.}
\end{table}

%% file: tables/app_01_defense_retrieval.tex
\begin{table*}[t] 
\resizebox{\textwidth}{!}{   
    \begin{tabular}{l|rrrrrrrrrrrrrrrr}
    \toprule
    $\lambda$ & arguana & bioasq & climate-fever & dbpedia-entity & fiqa & msmarco & nfcorpus & nq & quora & robust04 & scidocs & scifact & signal1m & trec-covid & trec-news & webis-touche2020 \\
    \midrule
 
    0 & 0.328 & 0.115 & 0.136 & 0.306 & 0.208 & 0.647 & 0.239 & 0.306 & 0.879 & 0.205 & 0.095 & 0.247 & 0.261 & 0.376 & 0.245 & 0.233 \\
    0.001 & 0.329 & 0.115 & 0.135 & 0.307 & 0.208 & 0.647 & 0.239 & 0.306 & 0.879 & 0.204 & 0.096 & 0.246 & 0.261 & 0.381 & 0.246 & 0.233 \\
    0.01 & 0.324 & 0.113 & 0.132 & 0.301 & 0.205 & 0.633 & 0.234 & 0.298 & 0.875 & 0.192 & 0.092 & 0.235 & 0.259 & 0.378 & 0.234 & 0.225 \\
    0.1 & 0.005 & 0.000 & 0.000 & 0.000 & 0.000 & 0.000 & 0.017 & 0.000 & 0.003 & 0.000 & 0.002 & 0.006 & 0.001 & 0.005 & 0.001 & 0.000 \\
    1.0 & 0.001 & 0.000 & 0.000 & 0.000 & 0.000 & 0.000 & 0.008 & 0.000 & 0.000 & 0.000 & 0.000 & 0.001 & 0.000 & 0.000 & 0.000 & 0.000 \\
    \bottomrule
    \end{tabular}
}
\label{app:tab:defense_retrieval_results}
\caption{BEIR performance (NDCG@10) for GTR-base at varying levels of noise (32 tokens). }
\end{table*}

%% file: tables/app_02_defense_agg.tex
\begin{table*}
\centering
\begin{tabular}{lrr}
\toprule
$\lambda$ & NDCG@10 & BLEU \\
\midrule
0.000 & 0.302 & 80.372 \\
0.001 & 0.302 & 72.347 \\
0.010 & 0.296 & 10.334 \\
0.100 & 0.002 & 0.148 \\
1.000 & 0.001 & 0.080 \\
\bottomrule
\end{tabular}
\label{app:tab:defense_aggregate_results}
\caption{Retrieval performance and reconstruction performance across varying noise levels $\lambda$.}
\end{table*}